\documentclass[10pt,twocolumn,letterpaper]{article}

\usepackage{cvpr}              
\usepackage[switch,mathlines]{lineno}

\definecolor{cvprblue}{rgb}{0.21,0.49,0.74}
\usepackage[pagebackref,breaklinks,colorlinks,allcolors=cvprblue]{hyperref}
\usepackage{pifont}
\usepackage[ruled,vlined]{algorithm2e}
\usepackage{booktabs}

\newcommand{\toolns}{\textit{ADvLM}}
\newcommand{\tool}{\toolns\space}

\newcommand{\Tref}[1]{Tab.~\ref{#1}}

\newcommand{\Fref}[1]{Fig.~\ref{#1}}
\newcommand{\Sref}[1]{Sec.~\ref{#1}}

\title{Visual Adversarial Attack on Vision-Language Models for \\Autonomous Driving}

\author{Tianyuan Zhang$^{1}$, Lu Wang$^{1}$, Xinwei Zhang$^{1}$, Yitong Zhang$^{1}$, Boyi Jia$^{1}$,\\ Siyuan Liang$^{2}$, Shengshan Hu$^{3}$, Qiang Fu$^{1}$, Aishan Liu$^{1}$, Xianglong Liu$^{1}$\\
$^{1}$Beihang University, China\\
$^{2}$National University of Singapore, Singapore\\
$^{3}$Huazhong University of Science and Technology, China\\
}

\begin{document}
\maketitle
\begin{abstract}
Vision-language models (VLMs) have significantly advanced autonomous driving (AD) by enhancing reasoning capabilities. However, these models remain highly vulnerable to adversarial attacks. While existing research has primarily focused on general VLM attacks, the development of attacks tailored to the safety-critical AD context has been largely overlooked. In this paper, we take the first step toward designing adversarial attacks specifically targeting VLMs in AD, exposing the substantial risks these attacks pose within this critical domain. We identify two unique challenges for effective adversarial attacks on AD VLMs: the variability of textual instructions and the time-series nature of visual scenarios. To this end, we propose \toolns, the first visual adversarial attack framework specifically designed for VLMs in AD. Our framework introduces Semantic-Invariant Induction, which uses a large language model to create a diverse prompt library of textual instructions with consistent semantic content, guided by semantic entropy. Building on this, we introduce Scenario-Associated Enhancement, an approach where attention mechanisms select key frames and perspectives within driving scenarios to optimize adversarial perturbations that generalize across the entire scenario. Extensive experiments on several AD VLMs over multiple benchmarks show that \tool achieves state-of-the-art attack effectiveness. Moreover, real-world attack studies further validate its applicability and potential in practice. 
\end{abstract}
\section{Introduction}
\label{sec:intro}

Owing to their strong generalization capabilities and inherent interpretability, vision-language models (VLMs) have demonstrated exceptional performance across various tasks, including autonomous driving (AD). By enabling autonomous systems to comprehend scenarios and process natural language, VLMs could serve as the brain and offer effective solutions for advanced reasoning in complex scenarios and more efficient human-machine interaction \cite{chen2024driving_with_llms, mao2023gptdriver, shao2024lmdrive, xu2024drivegpt4}. As a novel solution in end-to-end AD, VLMs present significant potential for future development.

\begin{figure}[!t]
\vspace{-0.1in}
\centering
\includegraphics[width=0.95\linewidth]{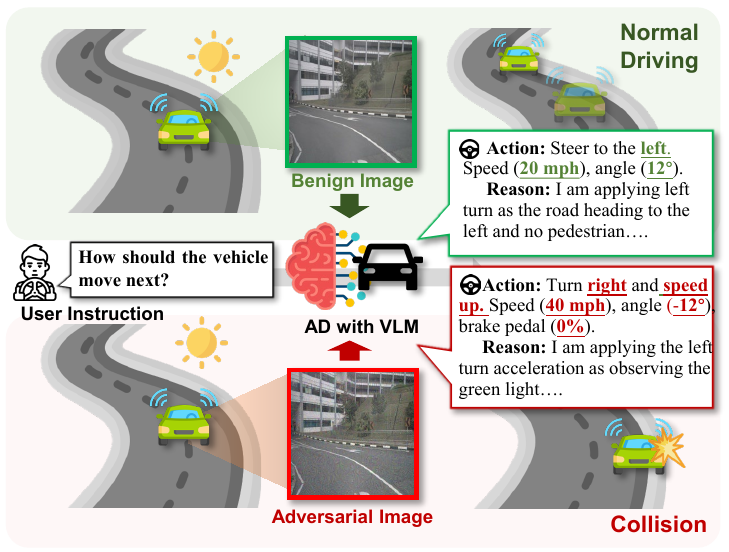}
\vspace{-0.15in}
\caption{Illustration of \toolns, where visual attacks lead model to generate incorrect decisions, demonstrating the potential risks associated with adversarial vulnerabilities in VLMs for AD.}
\label{fig:fontpage}
\vspace{-0.1in}
\end{figure}

However, VLMs exhibit significant vulnerabilities and lack robustness, particularly when faced with carefully crafted visual perturbations such as adversarial attacks \cite{ying2024safebench,ying2024jailbreak, yin2024vlattack, liang2021generate,liang2020efficient,wei2018transferable,liang2022parallel,liang2022large,wang2023diversifying,liu2023x,he2023generating,liu2023improving,lou2024hide,kong2024environmental,ma2021poisoning,ma2022tale,ma2024sequential, kong2024patch}. While various attack methods have been proposed, most existing research focuses on general VLMs and has not specifically addressed the unique requirements of AD. Identifying and addressing these vulnerabilities is essential in safety-critical domains like AD, as failures in VLMs could lead to severe consequences, including accidents or compromised decision-making.


In this paper, we take the first step to study adversarial attacks on VLM for AD. However, it is highly non-trivial to simply extend current adversarial attacks on general VLMs to this scenario, where We posit two key challenges unique in VLM for AD as follows. \ding{182} Attack should work among varied textual instructions with different phrases/sentences that convey the same task semantics. \ding{183} Attack should work for a specific time-series driving scenario with multiple visual frames and perspective shifts. To address these challenges, we propose \toolns, the first visual adversarial attack framework specifically tailored for VLMs in autonomous driving. In the textual modality, we propose the Semantic-Invariant Induction where we construct a low-semantic-entropy prompts library containing diverse textual instructions with the same semantics. Specifically, we employ a large language model to generate prompt variants from a seed and then refine them to promote the diversity in expressions guided by semantic entropy. In the visual modality, we introduce Scenario-Associated Enhancement, where we select critical frames/perspectives within the driving scenario based on model attentions, and further optimize the adversarial perturbations based on the pivotal frames while traversing the prompts library, such that the attack can generalize over the whole scenario. In this way, we can generate adversarial attacks across an expanded text and image input space, resulting in attacks that can remain effective and induce targeted behaviors across both varied instructions and time-series viewpoints in VLMs for AD.


To demonstrate its efficacy, we conduct extensive experiments on several VLMs for AD over multiple datasets, where our attack significantly outperforms other baselines with the highest Final Score reduction (+16.97\% and 7.49\%) in both white-box and black-box settings. In the closed-loop evaluation associated with the simulation environment CARLA, \tool also proves most effective, yielding a Vehicle Collisions Score of 2.954. In addition, we conduct real-world studies on physical vehicles to further demonstrate the potential of our attacks. Our \textbf{contributions} are shown as:

\begin{itemize}
\item We propose \toolns, the first adversarial attack specifically designed for VLMs in AD, addressing the unique challenges inherent in AD.

\item We introduce Semantic-Invariant Induction in the textual domain and Scenario-Associated Enhancement in the visual domain, ensuring attack effectiveness across varied instructions and sequential viewpoints.

\item Extensive experiments in both the digital and physical worlds demonstrate that \tool outperforms existing methods and shows high potential in practice.
\end{itemize}
\section{Related Works}
\label{sec:related works}

\textbf{Adversarial Attacks on VLMs}. With the widespread deployment and outstanding performance of VLMs in multimodal question answering and reasoning, their robustness~\cite{ying2024jailbreak, liu2023pre, liang2023badclip} has gradually attracted attention in recent years. Prior to our work, researchers have explored adversarial attacks against general VLMs. Due to the multimodal nature of VLMs, most adversarial attacks involve perturbations applied simultaneously to both image and text modalities. Drawing inspiration from adversarial attacks in vision tasks \cite{wang2021dual, liu2023x, li2023towards, liu2019perceptual, liu2020bias, zhang2021interpreting,liu2021training,liu2020spatiotemporal,liu2022harnessing,liu2023exploring,guo2023towards,liu2023towards, zhang2024lanevil, zhang2024towards}, these methods \cite{zhang2022towards, wang2024transferable, xu2024highly,zhang2023benchmarking,jiang2023exploring,jiang2024robuste2e,wang2024attack} typically rely on end-to-end differentiable gradients. \cite{zhang2022towards} introduced the first multimodal adversarial attack on VLMs, which paved the way for subsequent attacks that began exploring more practical black-box settings \cite{gao2025boosting, zhou2023advclip, yin2024vlattack}. Researchers typically aim for attack methods that introduce minimal perturbations while having a strong impact, leading some studies to focus solely on attacking the visual modality of VLMs. \cite{bailey2023image, zhang2024anyattack, zhao2024evaluating,zhang2024module} demonstrate that it is possible to attack specific targets using only image-based perturbations successfully. Adversarial attacks that target only the text modality are uncommon in VLMs, as they primarily focus on large language models.

Despite the development of various adversarial attack techniques for general VLMs, there is a notable lack of methods specifically addressing the robustness of VLMs in the safety-critical context of AD.

\textbf{VLMs in Autonomous Driving.} Recent research has increasingly focused on VLMs as a means to tackle AD tasks by integrating both visual and linguistic inputs. These models excel in tasks like perception, reasoning, and planning, which are essential for AD systems. The tasks of AD VLMs can be primarily categorized into two types. The first is the core function of VLMs, namely VQA, such as the classic Reason2Drive \cite{nie2023reason2drive}, LingoQA \cite{marcu2023lingoqa} and Dolphins \cite{ma2023dolphins}. These foundational works thoroughly explore the enhanced role of VLMs in AD, particularly their meticulous reasoning and explanatory abilities in various driving-related tasks such as scene understanding, behavior prediction, and dialogue. The second is driving planning or control, closely related to the operations of AD. GPT-Driver \cite{mao2023gptdriver}, Driving with LLMs \cite{chen2024driving_with_llms}, and MTD-GPT \cite{liu2023mtdgpt} pioneered improvements to VLMs for driving planning. However, these works only considered the driving problem in open-loop settings, overlooking issues such as cumulative errors and end-to-end interpretability. In contrast, LMDrive \cite{shao2024lmdrive} is the first to propose a VLM-based driving method within closed-loop settings, addressing these critical limitations. Other methods have integrated VQA and planning/control within VLM frameworks, offering a more holistic approach to AD. DriveLM \cite{sima2023drivelm}, DriveMLM \cite{wang2023drivemlm},m and DriveGPT4 \cite{xu2024drivegpt4} all go beyond basic conversations to implement more refined driving control and decision-making reasoning.

This paper selects representative models from each of the three categories for a comprehensive robustness analysis.




\section{Problem and Motivation}




\quad \textbf{Attack on VLMs.} Adversarial attacks on VLMs for AD aim to manipulate a model’s output by introducing carefully crafted perturbations into the input data. Specifically, an adversary applies adversarial perturbations to a benign query \((\mathbb{V}, t)\), where \(\mathbb{V} \subset \mathbb{V}_\text{all}\) denotes a specific sequence of visual inputs in AD, representing multiple frames rather than a single image. Here, \(\mathbb{V}_\text{all}\) is the domain of all possible visual inputs for the model. This results in an adversarial query \(\mathcal{P}(\mathbb{V}, t)\), where \(\mathcal{P}(\cdot)\) denotes the adversarial perturbation function. The goal of \(\mathcal{P}\) is to induce the VLM, denoted \(F_\theta\), to output a targeted or undesirable response \(y^*\) instead of the intended benign response. This manipulation is formally defined by maximizing the likelihood of the response \(y^*\) under the adversarial input:

\begin{equation}
\max_{\mathcal{P}} \log p(y^* | \mathcal{P}(\mathbb{V}, t)),
\end{equation}

\noindent where \(p\) represents the probability function \(F_\theta : \mathbb{Q} \rightarrow \mathbb{R}\), with \(\mathbb{Q} = \mathbb{V}_\text{all} \times \mathbb{T}\) as the input query domain, comprising sequences of visual inputs \(\mathbb{V} \subset \mathbb{V}_\text{all}\) and textual inputs \(t \in \mathbb{T}\), and \(\mathbb{R}\) as the response domain. In this work, we primarily focus on \emph{attacks in the visual domain}, ensuring generated perturbations remain consistent within the same sequence.

\begin{figure}[!t]
    \centering
    \includegraphics[width=0.95\linewidth]{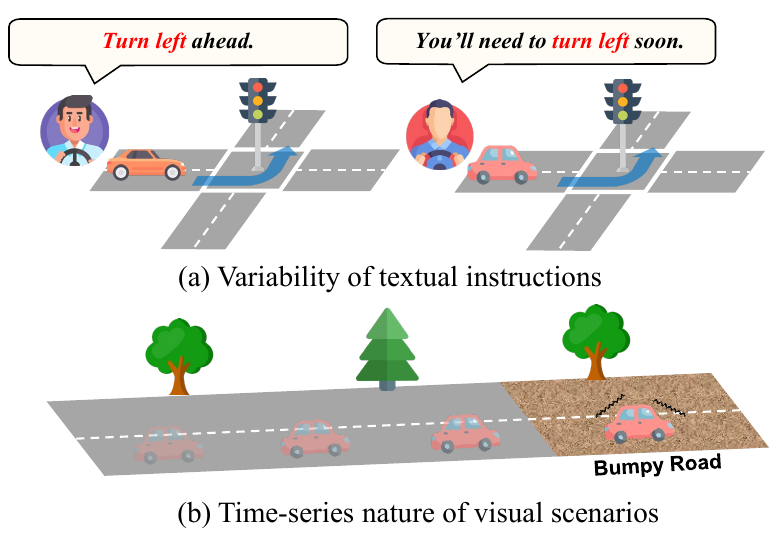}
    \vspace{-0.05in}
    \caption{Illustration of the main challenges of attacks for VLMs in AD including the variability of textual instructions and the time-series nature of visual scenarios.}
    \label{fig:challenges}
    \vspace{-0.1in}
\end{figure}

\textbf{Challenges and Attacking Goals.}
Common VLM adversarial attacks focus on fixed inputs (\ie, specific textual and visual input), but AD introduces unique challenges that require tailored approaches for effective attacks. We identify two key challenges essential for effective adversarial attacks in AD (as shown in \Fref{fig:challenges}), which differ this attack from those for general VLMs.

\ding{182} Variability of textual instructions. Drivers in AD often use varied textual instructions with different phrases for the same task, such as ``turn left at the intersection'' and ``turn left ahead''. In other words, these instructions are shown in different phrases but convey the same semantics and intent. To ensure a stable attack, visual perturbations \(\delta\) must remain effective across an expanded set of semantically equivalent prompts \(\widetilde{\mathbb{T}}\), derived from an original prompt, leading the VLM to consistently produce an incorrect response \(y^*\) across all prompts in \(\widetilde{\mathbb{T}}\) that convey the same command.


\ding{183} Time-series nature of visual scenarios. When driving, the vehicle’s perspective shifts frequently due to movement and environmental factors. AD models must adapt to visual changes from motion and temporal dependencies. Unlike static tasks, given an instruction, attacking AD VLMs demands the perturbations to make reliable impacts on a series of frames even as perspectives and image quality vary.  Let \(\widetilde{\mathbb{V}}\) represent a collection of different perspectives generated from an original frame in \(\mathbb{V}\), capturing the series of frames typical of time-series visual scenarios. This formulation ensures that the adversarial attack is effective across a dynamic visual sequence in an AD context.


To sum up, the adversary should consider generating adversarial perturbations that induce the AD VLM to produce the targeted response \( y^* \) consistently as follows:


\begin{equation}
\label{equ:challenge_total}
    \delta = {\arg \max} \sum_{{\mathbb{V}}} \sum_{\widetilde{\mathbb{T}}}\log p(y^* | \aleph(\widetilde{\mathbb{V}}, \delta), t),
\end{equation}

\noindent where \({t} \in \widetilde{\mathbb{T}} \) represents a specific prompt conveying the same semantic instruction, and \(\widetilde{\mathbb{V}}\subset \mathbb{V}\) indicates selected frames from the set of generated perspectives. The function \(\aleph(\widetilde{\mathbb{V}}, \delta)\) applies the perturbation \(\delta\) uniformly across all frames in \(\widetilde{\mathbb{V}}\). Optimizing \(\delta\) in this way ensures that the adversarial attack consistently misleads the model across varying perspectives and prompt formulations, thus enhancing robustness within the AD scenario.

\textbf{Threat Model.}
The adversary’s capabilities are limited to adding noise to image data, as interfering with the camera’s external inputs is easier than accessing the language module’s internal data. Given the sequential nature of AD, the adversary applies uniform noise $\delta_\star$ across the entire image sequence \(\mathbb{V}\), maintaining consistent perturbations within each sequence. The adversary’s knowledge differs by scenario, encompassing two primary AD threat models: white-box and black-box. In the white-box model, the adversary has full access to the model’s architecture, parameters, and data flow, allowing targeted exploitation of the model’s vulnerabilities. In contrast, the black-box model limits the adversary to indirect interactions, lacking insight into the model’s internal workings and requiring reliance on external observations. We assess \tool under these scenarios through open-loop and closed-loop experiments for the white-box setting (\emph{c.f.} \Sref{sec: open-loop}) and open-loop experiments for the black-box setting (\emph{c.f.} \Sref{black-box}).
\section{Approach}
\label{sec:method}

\begin{figure*}[!t]
\vspace{-0.2in}
    \centering
    \includegraphics[width=0.95\linewidth]{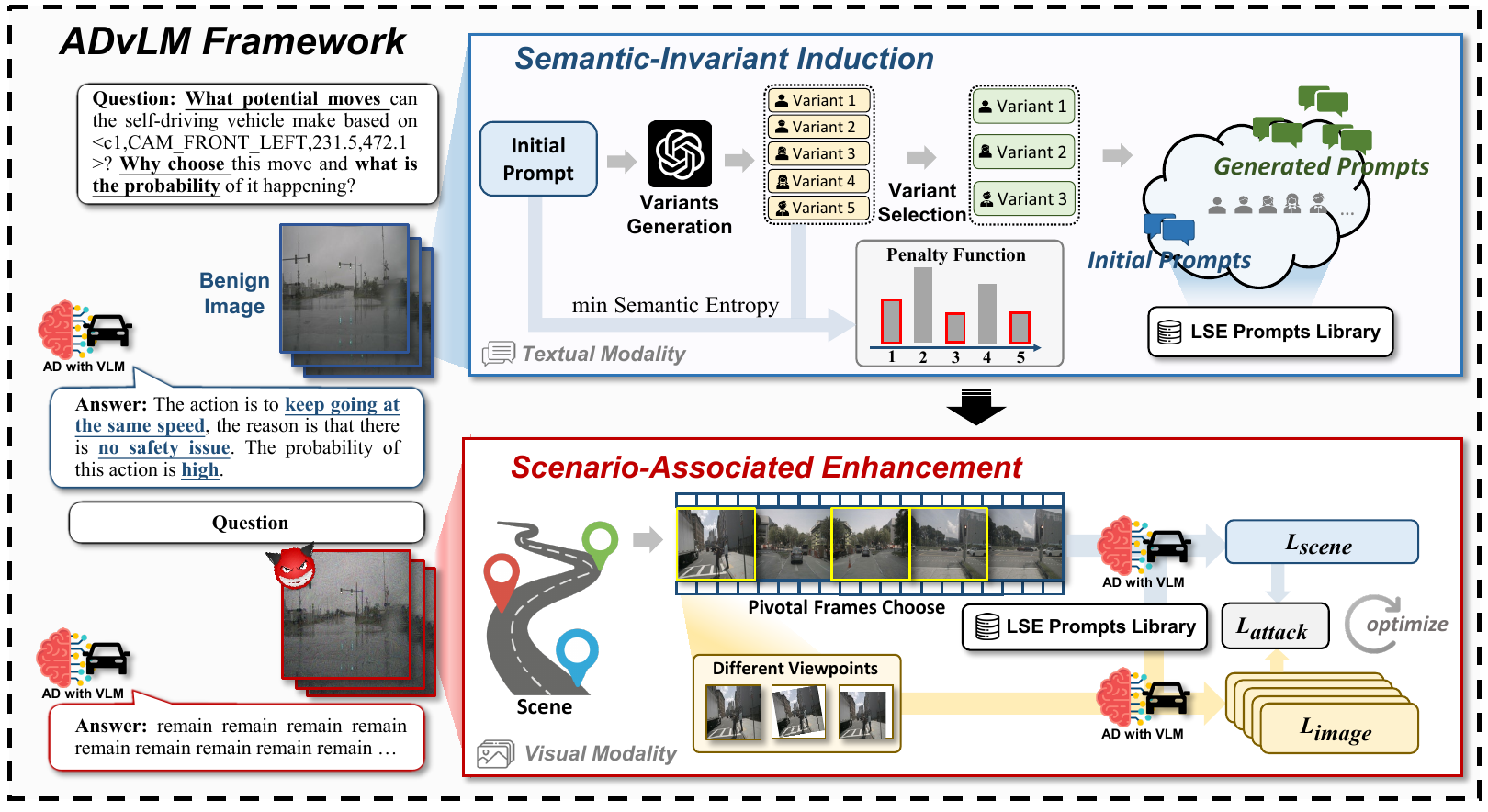}
    \vspace{-0.05in}
    \caption{The \tool Framework. \tool introduce Semantic-Invariant Induction in the textual domain and Scenario-Associated Enhancement in the visual domain, ensuring attack effectiveness across varied instructions and sequential viewpoints.}
    \label{fig: framework}
\vspace{-0.1in}
\end{figure*}
To address the above challenges, we propose \toolns, which exploits both the textual and visual modalities using proposed Semantic-Invariant Induction and Scenario-Associated Enhancement (as shown in \Fref{fig: framework}).


\subsection{Semantic-Invariant Induction}

In the textual modality, we introduce Semantic-Invariant Induction to construct a low-semantic-entropy (LSE) prompts library \( \widetilde{\mathbb{T}}_\text{LSE} \) containing diverse textual instructions with consistent semantic intent. Specifically, this approach leverages semantic entropy \cite{farquhar2024detecting_SE} to refine prompts generated from an initial seed, promoting expression diversity while retaining the same underlying meaning.

We employ GPT-4V \cite{achiam2023gpt} to generate semantically equivalent variants for each input \( t \). For each generated \( {t_i} \), we compute its semantic entropy \( \text{SE}({t_i}) \), aiming to achieve low entropy while enhancing expression variability. To guide this, we introduce a penalty function \( \mathcal{B}(t, {t_i}) \), balancing semantic consistency and expression diversity:

\begin{equation}
\label{equ:penalty_function}
\mathcal{B}(t, {t_i}) = \text{SE}({t_i}) + \beta \cdot \mathcal{D}(t, {t_i}),
\end{equation}

\noindent where \( \mathcal{D}(t, {t_i}) \) calculates expression similarity using Word2Vec embeddings \cite{church2017word2vec} and cosine similarity:

\begin{equation}
\mathcal{D}(t, {t_i}) = 1 - \frac{\Phi(t) \cdot \Phi({t_i})}{\|\Phi(t)\| \|\Phi({t_i})\|},
\end{equation}

\noindent with \(\Phi(x)\) representing the Word2Vec embedding of \( x \). The hyperparameter \(\beta\) controls the trade-off between entropy reduction and semantic alignment. The LSE prompt library for \( t \) is then defined as:

\begin{equation}
\label{equ: lse}
\widetilde{\mathbb{T}}_\text{LSE} = \bigcup_{t \in \mathbb{T}} \{{t_i} \ | \ \text{VLM}({t_i}), \ \min \ \mathcal{B}(t, {t_i})\}.
\end{equation}

This approach ensures expression diversity with minimized semantic entropy, creating a robust text modality to support effective adversarial attacks across varied AD instructions. 

\subsection{Scenario-Associated Enhancement}

In the visual modality, we introduce Scenario-Associated Enhancement (SAE) to enhance attack robustness across both textual instructions and visual frames in AD scenarios. Based on model attention, this method focuses on critical frames and perspectives identified within the driving scenario. The attack achieves generalization across the driving scenarios by refining adversarial perturbations for these pivotal frames while iterating through the LSE prompts library.

To ensure robustness across viewpoints, we design the image-wise loss \({L}_{\text{image}}\) with perspective transformations \(\mathcal{T}(\mathbb{V})\) applied to each visual input series \(\mathbb{V}\). This function ensures that the perturbations remain effective under diverse visual perspectives. The \({L}_{\text{image}}\) is defined as:

\begin{equation}
\label{equ:loss_image}
{L}_\text{image}(\mathbb{V}, t) = - \sum_{{t_i} \in \mathcal{S}(\widetilde{\mathbb{T}}_\text{LSE}, {t})} \log p(y^* | \mathcal{P}(\mathcal{T}(\mathbb{V}), {t_i}),
\end{equation}

\noindent where \(\widetilde{\mathbb{T}}_\text{LSE}\) represents the low-semantic-entropy prompt set, ensuring robustness across diverse textual inputs. The function \(\mathcal{S}(\cdot, \cdot)\) selects prompts with the same semantic meaning but different phrasings. 

To identify pivotal frames, we use an iterative attention-based selection process that maximizes diversity in attention maps across frames, enhancing scene coverage. Starting with the first frame \( {v_1} \) in sequence \( \mathbb{V} \) as the reference, we calculate the similarity between attention maps of each unselected frame \( {v_i} \) and the mean attention maps of the selected frames, using a similarity metric (average of SSIM \cite{wang2004image_SSIM} and PCC). For each candidate frame \( {v_i} \), we compute:

\begin{equation}
\label{equ: sim_calculation}
\text{Sim}({v_i}, \widetilde{\mathbb{V}}) =  \text{Sim}\left(\mathcal{A}({v_i}), \frac{1}{|\widetilde{\mathbb{V}}|} \sum_{{v} \in \widetilde{\mathbb{V}}} \mathcal{A}({v})\right),
\end{equation}

\noindent where \( \text{Sim}(\cdot, \cdot) \) measures similarity, and \( \widetilde{\mathbb{V}} \) is the set of selected frames. The next frame is chosen by minimizing the similarity to the set \( \widetilde{\mathbb{V}} \):

\begin{equation}
\label{equ: attention_choose}
{v_{\text{new}}} = {\arg \min}_{{v_i} \in \mathbb{V} \setminus \widetilde{\mathbb{V}}} \text{Sim}({v_i}, \widetilde{\mathbb{V}}).
\end{equation}

This selection continues until the desired number of frames $|\widetilde{\mathbb{V}}|$ is reached, ensuring each frame introduces distinct visual information.

Lastly, we apply a scene-wise loss \({L}_{\text{scene}}\) to optimize perturbations across these selected frames for enhanced generalization across varied environments:

\begin{equation}
\label{equ:loss_scene}
{L}_\text{scene} (\mathbb{V},{t}) = - \sum_{\widetilde{\mathbb{V}} \subset \mathbb{V}} \sum_{{t_i} \in \mathcal{S}(\widetilde{\mathbb{T}}_\text{LSE}, {t})} \log p(y^* | \mathcal{P}(\widetilde{\mathbb{V}}, {t_i})).
\end{equation}


\subsection{Overall Attack Process}

The primary objective of this attack is to minimize the loss \({L}_\text{attack}\), ensuring that the perturbation remains effective despite variations in both text and perspective, thereby expanding the adversarial space and enhancing robustness. The combined loss function is defined as:

\begin{equation}
\label{equ:loss_total}
{L}_\text{attack} = (1 - \lambda) \cdot {L}_{\text{image}} + \lambda \cdot {L}_{\text{scene}},
\end{equation}

\noindent where \(\lambda\) controls the contribution of the \({L}_{\text{scene}}\). To balance the influence between image-wise and scene-wise losses, we set the hyper-parameter \(\lambda\) to 0.4. 




    
    
    

        
        
    
\section{Experiments}
\label{sec: experiments}


\subsection{Experimental Settings}
\label{sec: exp-setup}

\quad \textbf{Target Models.}
We select 3 state-of-the-art VLM-based AD models for attack including DriveLM \cite{sima2023drivelm}, Dolphins \cite{ma2023dolphins}, and LMDrive \cite{shao2024lmdrive}. In addition, we also evaluate our attacks on 4 general VLMs including MiniGPT-4 \cite{zhu2023minigpt}, MMGPT \cite{gong2023multimodal}, LLaVA \cite{liu2024visual}, and GPT-4V \cite{achiam2023gpt}.

\textbf{Evaluation Datasets.}
We evaluate our approach under both open-loop and closed-loop settings. For open-loop conditions, we use the DriveLM-\tool and Dolphins-\tool datasets, which are expanded from Drivelm-nuScenes \cite{sima2023drivelm} and Dolphins Benchmark \cite{ma2023dolphins}. For closed-loop conditions, we use the LangAuto-Tiny benchmark \cite{shao2024lmdrive} scenarios, and CARLA simulators generate the input data based on these scenarios.

\textbf{Evaluation Metrics.}  
For DriveLM and Dolphins, we calculate a weighted average of language metrics and GPT-Score, following the approach in \cite{sima2023drivelm} and \cite{ma2023dolphins}. For closed-loop conditions, we use metrics provided by the CARLA leaderboard \cite{dosovitskiy2017carla}. Given that linguistic quality is less critical in AD systems, we reduced the weight of the Language Score and adjusted the other metrics to create a New Final Score in the evaluation of DriveLM. \textit{$\textcolor{blue}{\downarrow}$ indicates the lower the better attack, while $\textcolor{red}{\uparrow}$ indicates higher the better.}


\textbf{Attack baselines}. We choose 2 classical adversarial attacks including FGSM \cite{FGSM}, PGD \cite{PGD}, and 2 commonly adopted attacks on VLMs (AttackVLM \cite{zhao2024evaluating}, and AnyAttack \cite{zhang2024anyattack}) for comparison.

\textbf{Implementation Details.}
For our \toolns, we empirically set $\lambda = 0.4$, with $\epsilon = 0.1$, $n = 50$ and $\alpha = 2 * \epsilon / n$. All code is implemented in PyTorch, and experiments are conducted on an NVIDIA A800-SXM4-80GB GPU cluster. 

\textit{More details about our experimental settings can be found in the Supplementary Material.}





\subsection{White-box Attack}
We first perform white-box attacks in both the open-loop (static, controlled environment with predefined inputs) and closed-loop scenarios (dynamic, interactive environment with real-time feedback and model adaptation).

\textbf{Open-loop Evaluation}. For the number of pivotal frames, we set \( | \widetilde{\mathbb{V}} | = 6 \) for the Dolphins model, which processes video frames as input. For DriveLM, which operates on single images rather than consecutive frames, we use \( | \widetilde{\mathbb{V}} | = 1 \). The attack results are presented in \Tref{tab: open-loop}, leading to the following observations.
\label{sec: open-loop}

\ding{182} Our \tool method achieves significantly better performance on different models (a maximum final score drop by 16.97\% on DriveLM and 9.64\% on Dolphins). 

\ding{183} We observed that AttackVLM and AnyAttack perform comparatively worse than other baselines. We hypothesize that this may be due to these methods being primarily designed for black-box attacks, leading to lower effectiveness in white-box settings. Therefore, we conduct additional black-box attack experiments in \Sref{black-box}.

\ding{184} In the evaluation of Dolphins, the performance of \tool on Time tasks is slightly lower than that of PGD. Detailed experiments indicate that adjusting the hyperparameter \(\lambda\) can effectively enhance performance on the Time task. For more information, please refer to \Sref{ablation}.



\ding{185} Notably, in the evaluation of DriveLM, \tool reduces the Language Score by 13.20\%, which is less than the 17.96\% drop achieved by the PGD method. This does not indicate weaker attack effectiveness; rather, since the Language Score reflects linguistic quality, a higher score can make it harder for drivers to detect the attack, potentially delaying their intervention. \textit{We provided a detailed explanation in the Supplementary Material.}

\begin{table}[!t]
\centering
\renewcommand\arraystretch{0.8}
\caption{Evaluation results under open-loop conditions. \textbf{Bold text} indicates the method with the strongest attack effect in each column. \colorbox[gray]{0.9}{Gray cells} represent comprehensive evaluation metrics.}
\vspace{-0.05in}
\small
\label{tab: open-loop}
\subfloat[DriveLM]{
\label{tab:drivelm}
\resizebox{\linewidth}{!}{
\begin{tabular}{@{}c|cccccc@{}}
\toprule
\textbf{Method/Metrics} & Accuracy$\textcolor{blue}{\downarrow}$ & Chatgpt$\textcolor{blue}{\downarrow}$ & Match$\textcolor{blue}{\downarrow}$ & Language$\textcolor{blue}{\downarrow}$ & Final$\textcolor{blue}{\downarrow}$ & Final$^\dagger$$\textcolor{blue}{\downarrow}$ \\ \midrule
Raw                                  & 71.43 & 66.60 & 31.73 & 46.39 & \cellcolor[HTML]{EFEFEF}56.55 & \cellcolor[HTML]{EFEFEF}53.62 \\

FGSM \cite{FGSM}                     & 73.81 & 67.26 & 32.28 & 39.44 & \cellcolor[HTML]{EFEFEF}56.01 & \cellcolor[HTML]{EFEFEF}54.58 \\

PGD \cite{PGD}                       & 61.90 & 48.45 & 25.20 & \textbf{28.43} & \cellcolor[HTML]{EFEFEF}42.49 & \cellcolor[HTML]{EFEFEF}41.84 \\

AttackVLM \cite{zhao2024evaluating}  & 70.12 & 63.81 & 30.15 & 42.50 & \cellcolor[HTML]{EFEFEF}54.08 & \cellcolor[HTML]{EFEFEF}51.61 \\

AnyAttack \cite{zhang2024anyattack}  & 71.20 & 64.05 & 30.95 & 43.10 & \cellcolor[HTML]{EFEFEF}54.67 & \cellcolor[HTML]{EFEFEF}52.24 \\ \midrule

\toolns & \textbf{52.38} & \textbf{43.73} & \textbf{24.86} & 33.19 & \cellcolor[HTML]{EFEFEF}\textbf{39.58} & \cellcolor[HTML]{EFEFEF}\textbf{37.92} \\ \bottomrule
\end{tabular}

}
}

\raggedright
\footnotesize{$^\dagger$ New Final Score.}

\vspace{0.1cm}
\subfloat[Dolphins]{
\label{tab:dolphins}
\resizebox{\linewidth}{!}{
\begin{tabular}{@{}c|ccccccc@{}}
\toprule[0.75pt]
\textbf{Method/Metrics}                   & Weather$\textcolor{blue}{\downarrow}$ & Traffic.$\textcolor{blue}{\downarrow}$ & Time$\textcolor{blue}{\downarrow}$ & Scene$\textcolor{blue}{\downarrow}$ & Open.$\textcolor{blue}{\downarrow}$ & Desc$\textcolor{blue}{\downarrow}$ & Final$\textcolor{blue}{\downarrow}$ \\ \midrule
Raw & 48.75 & 52.82 & 39.71 & 43.31 & 29.79 & 41.37 & \cellcolor[HTML]{EFEFEF}42.63 \\

FGSM \cite{FGSM} & 47.27 & 52.99 & 46.62 & 45.46 & 23.60 & 45.71 & \cellcolor[HTML]{EFEFEF}43.61 \\

PGD \cite{PGD} & 32.43 & 40.67 & \textbf{36.51} & 33.04 & 22.15 & 36.83 & \cellcolor[HTML]{EFEFEF}33.60 \\

AttackVLM \cite{zhao2024evaluating} & 44.32 & 50.60 & 41.12 & 43.25 & 28.98 & 42.51 & \cellcolor[HTML]{EFEFEF}41.96 \\

AnyAttack \cite{zhang2024anyattack} & 45.10 & 51.12 & 43.28 & 44.10 & 27.25 & 43.14 & \cellcolor[HTML]{EFEFEF}42.28 \\ \midrule
\toolns & \textbf{31.09} & \textbf{41.06} &  39.36 & \textbf{32.60} & \textbf{17.44} & \textbf{36.45} & \cellcolor[HTML]{EFEFEF}\textbf{32.99} \\ \bottomrule[0.75pt]
\end{tabular}
}
}
\vspace{-0.1in}
\end{table}

\textbf{Closed-loop Evaluation}.
\label{sec: closed-loop}
For the closed-loop evaluation, we used the pre-trained model provided by LMDrive \cite{shao2024lmdrive}. Since LMDrive operates on single images rather than consecutive frames, we set \( | \widetilde{\mathbb{V}} | = 1 \). The evaluation pipeline follows these steps: \ding{182} start the Docker version of CARLA 0.9.10.1, \ding{183} launch the CARLA leaderboard with a specified agent, and \ding{184} activate drive mode and begin the evaluation. Due to variability in traffic flow and decision-making, the results can be unstable; therefore, we averaged the results over multiple trials. Each experimental setting was run five times, with metrics reported as the average of these repetitions. The evaluation results are shown in \Tref{tab: closed-loop}.

Our \tool method outperforms all other attack methods, achieving a 23.88\% reduction in infraction penalty, with increases in collisions with vehicles and layout. Notably, the performance of \tool on off-road infractions is within 0.5\% lower than PGD, likely due to the higher sensitivity of PGD to specific boundary conditions. However, this is a minor difference compared to the overall improvements achieved by \tool across other metrics.

\begin{figure}[t]
\centering
        \includegraphics[width=0.93\linewidth]{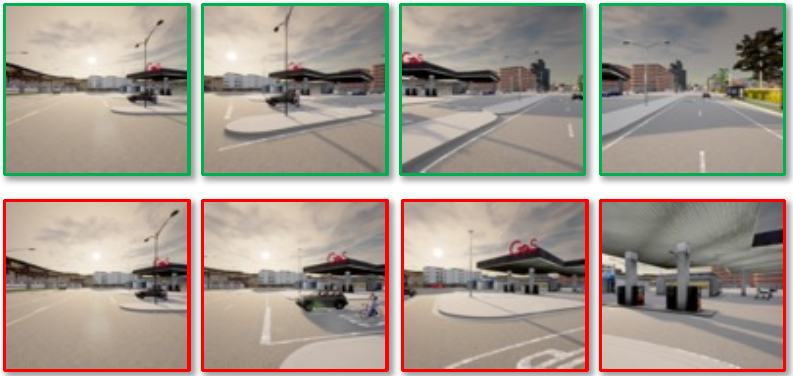}
\caption{Closed-loop exp in Town 03 Route 26 of CARLA. After the attack (\textcolor{green}{Green} $\rightarrow$ \textcolor{red}{Red}), the vehicle veers towards the gas station, highlighting real-world potential safety risks.}
\label{fig:route}
\end{figure}

Additionally, we present visualizations from the experiment on Town 03 Route 26 in \Fref{fig:route}. Before the attack, the vehicle navigated normally; however, after the attack, it veered into a gas station, posing a significant safety risk and underscoring potential security vulnerabilities.

\begin{table}[!t]
\centering
\renewcommand\arraystretch{0.9}
\caption{Evaluation Results under closed-loop conditions. \textbf{Bold text} indicates the most effective attack in each column.}
\label{tab: closed-loop}
\resizebox{\linewidth}{!}{
\begin{tabular}{@{}c|ccccc@{}}
\toprule
\textbf{Method/Metrics}        & \multicolumn{1}{l}{Infraction.$\textcolor{blue}{\downarrow}$} & \multicolumn{1}{l}{Vehicle.$\textcolor{red}{\uparrow}$} & \multicolumn{1}{l}{Layout.$\textcolor{red}{\uparrow}$} & \multicolumn{1}{l}{Red lights.$\textcolor{red}{\uparrow}$} & \multicolumn{1}{l}{Off-road.$\textcolor{red}{\uparrow}$} \\ \midrule
Raw & 0.787 & 0.832 & 1.989 & 0.654 & 1.437 \\

FGSM \cite{FGSM} & 0.721 & 0.663 & 4.494 & 0.785 & 4.577 \\

PGD \cite{PGD} & 0.608 & 1.454 & 5.704 & 1.321 & \textbf{5.203} \\

AttackVLM \cite{zhao2024evaluating} & 0.775 & 0.810 & 1.950 & 0.670 & 1.450 \\

AnyAttack \cite{zhang2024anyattack} & 0.780 & 0.820 & 1.980 & 0.660 & 1.420 \\ \midrule

\toolns & \textbf{0.599} & \textbf{2.954} & \textbf{6.869} & \textbf{1.473} & 5.188 \\ \bottomrule
\end{tabular}
}\end{table}

\subsection{Black-box Evaluation}
\label{black-box}

\quad \textbf{Black-box Settings.}
In contrast to the white-box setting, where an adversary has full access to model details, the black-box scenario limits the attacker to model input/output, without insight into the model’s internal structure. Our black-box evaluation is conducted in open-loop experiments, where we adapt the models and datasets of DriveLM \cite{sima2023drivelm} and Dolphins \cite{ma2023dolphins} in novel ways to enable transfer-based attacks. Specifically, we use Dolphins as the victim model with DriveLM as the substitute model, applying the Dolphins-\tool dataset and white-box Dolphins for attack generation and performing attacks on DriveLM. The same approach is used for DriveLM in the black-box setting. We employ transfer-based methods for \toolns, FGSM, and PGD, while directly implementing AttackVLM \cite{zhao2024evaluating} and AnyAttack \cite{zhang2024anyattack}, as these methods are inherently designed for black-box environments.

\textbf{Results Analysis.} 
The black-box evaluation results are provided in \Tref{tab:drivelm-b} and \Tref{tab:dolphins-b}, using the same metrics as outlined in \Sref{sec: exp-setup}. The findings reveal that across both DriveLM and Dolphins models, \tool consistently achieves lower Final Scores than other methods, with reductions of up to 7.49\% on DriveLM and 3.09\% on Dolphins. This significant performance decline across varied datasets demonstrates the high effectiveness of \tool in degrading model performance in black-box settings, establishing it as a robust approach for transfer-based adversarial attacks.

\textbf{Attack on General VLMs.} We also conducted experiments on general VLMs (\ie, MiniGPT-4 \cite{zhu2023minigpt}, MMGPT \cite{gong2023multimodal}, LLaVA \cite{liu2024visual}, and GPT-4V \cite{achiam2023gpt}) using DriveLM-\toolns with attack noise generated from DriveLM. Results, shown in \Tref{tab:general} and measured by Final Score$\textcolor{blue}{\downarrow}$, reveal that while general-purpose models perform acceptably in AD tasks, there is a substantial performance gap compared to VLMs specifically designed for AD. In terms of attack effectiveness, \toolns, AttackVLM, and AnyAttack exhibit the strongest impact, demonstrating that our method effectively compromises general VLMs as well.

\begin{table}[!t]
\centering
\renewcommand\arraystretch{0.8}
\caption{Evaluation results under black-box settings. \textbf{Bold text} indicates the method with the strongest attack effect in each column. \colorbox[gray]{0.9}{Gray cells} represent comprehensive evaluation metrics.}
\small
\label{tab:black}
\subfloat[Results on DriveLM-\tool use DriveLM: Transfer from Dolphins]{
\label{tab:drivelm-b}
\resizebox{0.95\linewidth}{!}{
\begin{tabular}{@{}c|cccccc@{}}
\toprule
\textbf{Method/Metrics} & Accuracy$\textcolor{blue}{\downarrow}$ & Chatgpt$\textcolor{blue}{\downarrow}$ & Match$\textcolor{blue}{\downarrow}$ & Language$\textcolor{blue}{\downarrow}$ & Final$\textcolor{blue}{\downarrow}$ & Final$^\dagger$$\textcolor{blue}{\downarrow}$ \\ \midrule
Raw                                  & 71.43 & 66.60 & 31.73 & 46.39 & \cellcolor[HTML]{EFEFEF}56.55 & \cellcolor[HTML]{EFEFEF}53.62 \\

FGSM \cite{FGSM}                & 69.61    & 60.33 & 27.29 &  46.12 & \cellcolor[HTML]{EFEFEF}51.74 & \cellcolor[HTML]{EFEFEF}48.97 \\

PGD \cite{PGD}                       & 69.44 & 61.07 & 27.19 & \textbf{41.74} & \cellcolor[HTML]{EFEFEF}52.10 & \cellcolor[HTML]{EFEFEF}49.19 \\

AttackVLM \cite{zhao2024evaluating}  & 70.12 & 63.81 & 30.15 & 42.50 & \cellcolor[HTML]{EFEFEF}54.08 & \cellcolor[HTML]{EFEFEF}51.61 \\

AnyAttack \cite{zhang2024anyattack}  & 71.20 & 64.05 & 30.95 & 43.10 & \cellcolor[HTML]{EFEFEF}54.67 & \cellcolor[HTML]{EFEFEF}52.24 \\ \midrule

\toolns & \textbf{66.25}  & \textbf{56.24} & \textbf{23.91} & 42.64 & \cellcolor[HTML]{EFEFEF}\textbf{49.06} & \cellcolor[HTML]{EFEFEF}\textbf{45.31} \\ \bottomrule
\end{tabular}

}
}

\raggedright
\footnotesize{$^\dagger$ New Final Score.}

\vspace{0.2cm}
\subfloat[Results on Dolphins-\tool use Dolphins: Transfer from DriveLM]{
\label{tab:dolphins-b}
\resizebox{0.95\linewidth}{!}{
\begin{tabular}{@{}c|ccccccc@{}}
\toprule[0.75pt]
\textbf{Method/Metrics}                   & Weather$\textcolor{blue}{\downarrow}$ & Traffic.$\textcolor{blue}{\downarrow}$ & Time$\textcolor{blue}{\downarrow}$ & Scene$\textcolor{blue}{\downarrow}$ & Open.$\textcolor{blue}{\downarrow}$ & Desc$\textcolor{blue}{\downarrow}$ & Final$\textcolor{blue}{\downarrow}$ \\ \midrule
Raw & 48.75 & 52.82 & 39.71 & 43.31 & 29.79 & 41.37 & \cellcolor[HTML]{EFEFEF}42.63 \\

FGSM \cite{FGSM} & 48.62 & 51.34 & 38.74 & 40.16 & 27.00 & 41.10 & \cellcolor[HTML]{EFEFEF}41.16 \\

PGD \cite{PGD} & 48.01 & 52.79 & \textbf{38.46} & 38.58 & \textbf{24.87} & 40.46 & \cellcolor[HTML]{EFEFEF}40.53 \\

AttackVLM \cite{zhao2024evaluating} & 44.32 & 50.60 & 41.12 & 43.25 & 28.98 & 42.51 & \cellcolor[HTML]{EFEFEF}41.96 \\

AnyAttack \cite{zhang2024anyattack} & 45.10 & 51.12 & 43.28 & 44.10 & 27.25 & 43.14 & \cellcolor[HTML]{EFEFEF}42.28 \\ \midrule
\toolns & \textbf{43.83} & \textbf{49.98} &  39.17 & \textbf{36.17} & 28.70 & \textbf{39.44} & \cellcolor[HTML]{EFEFEF}\textbf{39.54} \\ \bottomrule[0.75pt]
\end{tabular}
}}

\vspace{0.2cm}
\subfloat[Results on DriveLM-\tool use VLMs: Transfer from DriveLM]{
\label{tab:general}
\large
\resizebox{\linewidth}{!}{
\begin{tabular}{@{}c|ccccc@{}}
\toprule[1.0pt]
\textbf{Method/Model} & MiniGPT-4 \cite{zhu2023minigpt} & MMGPT \cite{gong2023multimodal} & LLaVA \cite{liu2024visual} & GPT-4V \cite{achiam2023gpt}  \\ \midrule
Raw & 46.23 & 45.58 & 48.77 & 49.89  \\

FGSM \cite{FGSM} & 41.12 & 43.83 & 42.18 & 44.63 \\

PGD \cite{PGD} & 35.47 & 41.28 & 37.82 & 39.54 \\

AttackVLM \cite{zhao2024evaluating} & \textbf{30.13} & 38.21 & 33.57 & 36.92 \\

AnyAttack \cite{zhang2024anyattack} & 31.93 & 39.02 & 31.23 & \textbf{34.83} \\ \midrule
\toolns & 30.52 & \textbf{37.43} & \textbf{30.83} & 35.13 \\ \bottomrule[1.0pt]
\end{tabular}

}}
\end{table}

\subsection{Ablation Studies}
\label{ablation}

\begin{figure}[t]
\centering
	\begin{subfigure}{0.48\linewidth}
		\centering
        \includegraphics[width=0.95\linewidth]{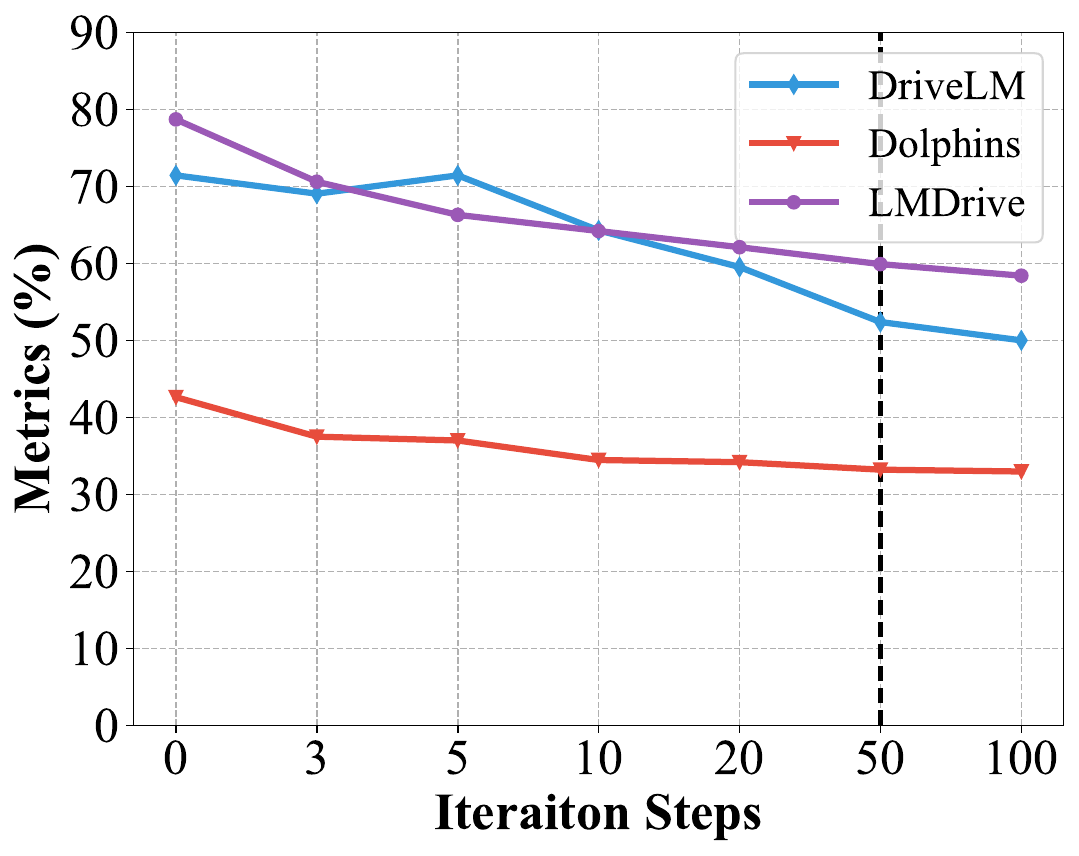}
		\caption{Different Steps}
		\label{subfig:different_steps}
	\end{subfigure}
	\begin{subfigure}{0.48\linewidth}
		\centering
		\includegraphics[width=0.95\linewidth]{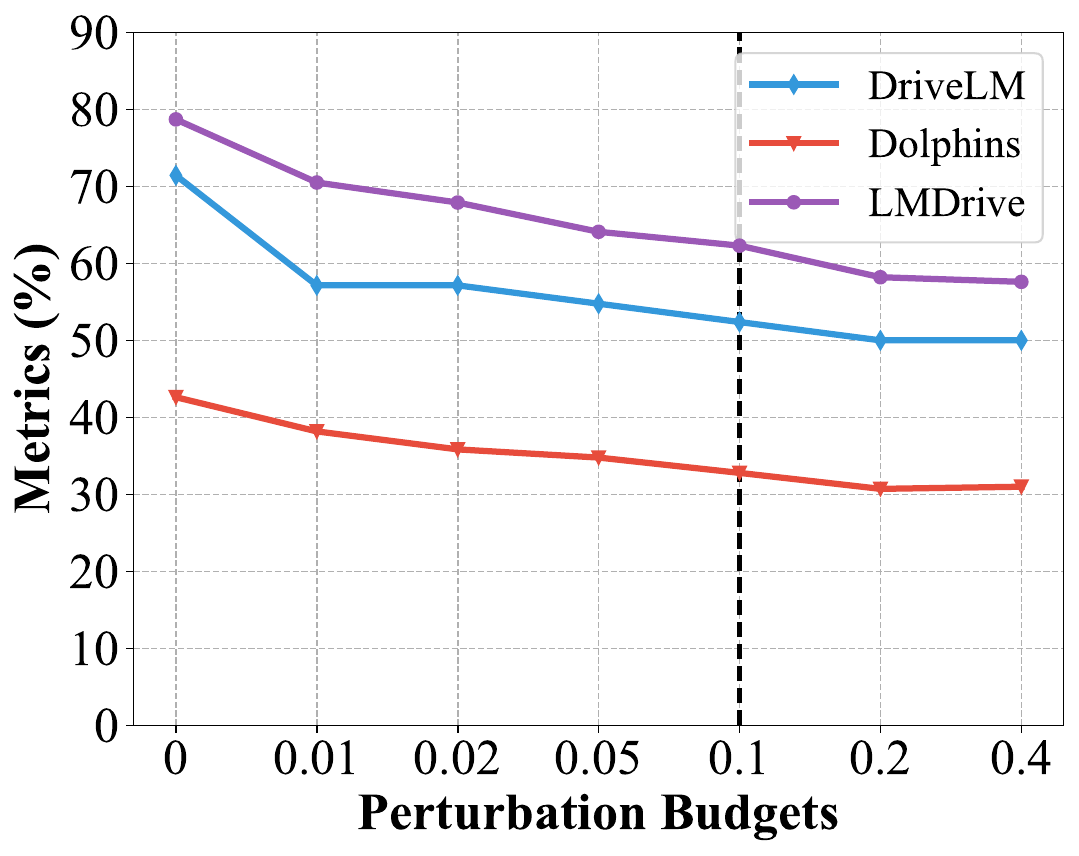}
		\caption{Different Budgets}
		\label{subfig:different_budgets}
	\end{subfigure}
\vspace{-0.05in}
\caption{Experiments results under different steps and budgets. The main experiment settings are marked with black dashed lines.}
\label{fig:zhexian}
\vspace{-0.1in}
\end{figure}

\quad \textbf{Perturbation Budgets and Step Sizes.}
We conducted an ablation study to explore the impact of different attack settings. First, we present the results of \tool attacks with varying iteration steps \(n\) (\ie, 3, 5, 10, 20, 50, 100) on DriveLM-\tool and Dolphins-\toolns, using a fixed perturbation budget of \(\epsilon=0.1\) and step size \(\alpha=2\epsilon/n\). Generally, the attack strength increases with more iteration steps, as shown in \Fref{subfig:different_steps}. Additionally, we tested \tool with different perturbation budgets \(\epsilon\) (\ie, 0.01, 0.02, 0.05, 0.1, 0.2, 0.4) across three models, with \(n=50\) and \(\alpha=2\epsilon/n\). The specific budgets and results are shown in \Fref{subfig:different_budgets}. For DriveLM and Dolphins, we evaluate performance using the Final Score$\textcolor{blue}{\downarrow}$, while for LMDrive, we use the Infraction Score$\textcolor{blue}{\downarrow}$. The results indicate that as \(n\) and \(\epsilon\) increase, attack effectiveness improves but levels off when \(n=50\) and \(\epsilon=0.1\). Therefore, we selected these values.

\begin{figure}[!t]
\centering
	\begin{subfigure}{0.48\linewidth}
		\centering
        \includegraphics[width=0.95\linewidth]{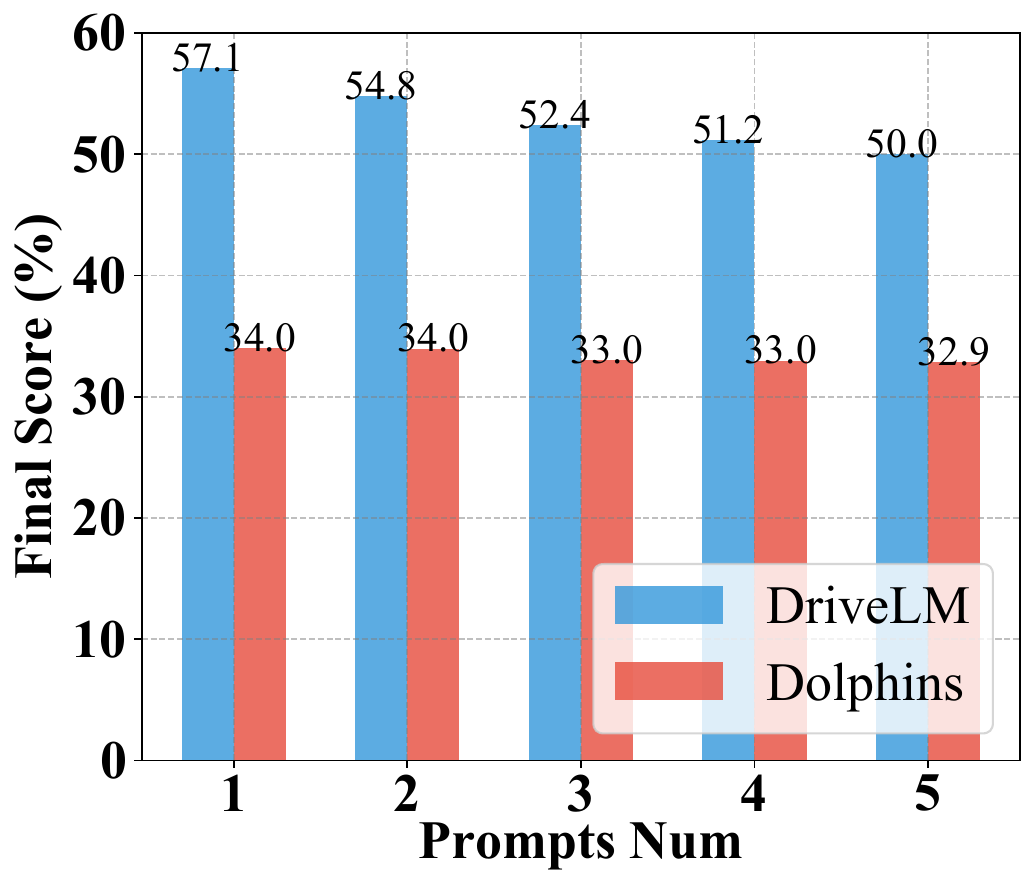}
		\caption{With SAE.}
		\label{subfig:different_prompts}
	\end{subfigure}
	\begin{subfigure}{0.48\linewidth}
		\centering
		\includegraphics[width=0.95\linewidth]{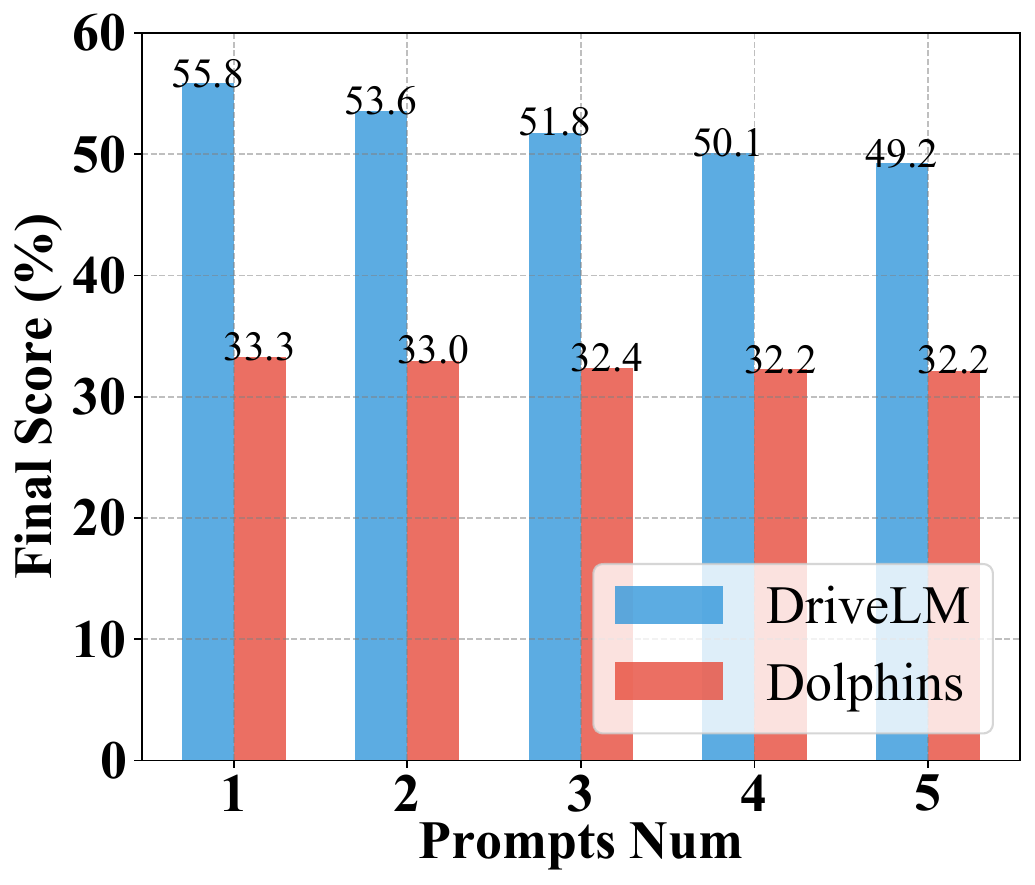}
		\caption{Without SAE.}
		\label{subfig:different_prompts_without_variable}
	\end{subfigure}
\caption{Results under different numbers of prompts.}
\label{fig:zhuzhuang}
\vspace{-0.1in}
\end{figure}

\textbf{Semantic-Invariant Induction.}
We conducted experiments with different numbers of prompts (\ie, 1, 2, 3, 4, and 5), using iteration steps \(n=5\) and \(\epsilon=0.1\). Results are shown in \Fref{subfig:different_prompts}. As the number of prompts increases, attack effectiveness improves. When prompts are increased from 1 to 3, the Final Score$\textcolor{blue}{\downarrow}$ of DriveLM and Dolphins decreases from 57.14\% and 33.99\% to 52.38\% and 33.03\%, respectively. However, this improvement becomes marginal beyond 3 prompts, with accuracy only slightly decreasing to 50.0\% and 32.88\% at 5 prompts. We believe that three LSE prompts sufficiently capture the semantic information needed for effective attacks.

\textbf{Series-Associated Enhancement.}  
We conducted experiments without the variable perspective technique, using the same setup as described previously but omitting the variable perspective method. Results are shown in \Fref{subfig:different_budgets}. The data shows a similar trend but with an average increase of 2.12\% compared to the previous experiment. The experimental results validated the effectiveness of the variable perspective.

\textbf{Hyper-parameter $\lambda$.} 
We evaluate the effect of \(\lambda\) on Dolphins using the Final Score$\textcolor{blue}{\downarrow}$, varying \(\lambda\) from 0.1 to 0.9 in steps of 0.1. 
Optimal attack performance occurs at \(\lambda = 0.4\), though certain tasks, like Time, peak at \(\lambda = 0.6\). This sensitivity to \(\lambda\) highlights \(\lambda\)'s role in tuning adversarial impact across tasks.

\textbf{The number of pivotal frames $|\widetilde{\mathbb{V}}|$.}
We assess the influence of \( |\widetilde{\mathbb{V}}| \) on Dolphins using the Final Score\(\textcolor{blue}{\downarrow}\), adjusting \( |\widetilde{\mathbb{V}}| \) from 1 to 16 in increments of 1, as the longest scene in Dolphins consists of 16 frames per prompt. Results show that the optimal attack performance is achieved at \( |\widetilde{\mathbb{V}}| = 6 \) with 39.54, the lowest observed in the experiments. For other values, we observe less effective performance, such as 42.31 at \( |\widetilde{\mathbb{V}}| = 4 \) and 41.67 at \( |\widetilde{\mathbb{V}}| = 8 \), underscoring that 6 frames offer a balanced yet effective representation for inducing the most robust adversarial impact.

\subsection{Discussion and Analysis}

\quad \textbf{Analysis of Textual Instruction Variability.}
We conduct experiments to assess the impact of textual variability on attack effectiveness. Using the DriveLM-\toolns dataset, which includes both standard prompts and sets of expanded, semantically equivalent prompts, we evaluate four attack methods (\ie, \toolns, PGD, AnyAttack, and AttackVLM) under varied textual conditions. The evaluation metric is Final Score$\textcolor{blue}{\downarrow}$, with the extended dataset tested by expanding each test case to include 3 and 5 semantically similar prompts, respectively, and calculating the final result as their average. As shown in \Tref{tab:analysis}, \tool consistently maintains high attack effectiveness across the expanded dataset, while other methods experience notable declines in performance as textual variability increases.

\begin{table}[!t]

\caption{Analysis of textual instruction variability. Values in \textcolor{blue}{blue} indicate the reduction relative to Raw.}
\large
\label{tab:analysis}
\resizebox{\linewidth}{!}{
\begin{tabular}{@{}cccccc@{}}
\toprule
\textbf{Datasets/Method} & Raw  & PGD \cite{PGD} & AttackVLM \cite{zhao2024evaluating} & AnyAttack \cite{zhang2024anyattack} & \toolns \\ \midrule
DriveLM-\toolns & 56.55 & 42.49 / \textcolor{blue}{14.06} & 54.08 / \textcolor{blue}{2.47} & 54.67 / \textcolor{blue}{1.88} & 39.58 / \textcolor{blue}{\textbf{16.97}}  \\
DriveLM-\toolns$^\dagger$ & 56.79 & 49.14 / \textcolor{blue}{7.65} & 54.78 / \textcolor{blue}{2.01} & 55.14 / \textcolor{blue}{1.65} & 44.78 / \textcolor{blue}{\textbf{11.01}}   \\
DriveLM-\toolns$^\ddagger$ & 57.16 & 55.49 / \textcolor{blue}{1.67} & 55.35 / \textcolor{blue}{1.81} & 55.49 / \textcolor{blue}{1.67} & 50.54 / \textcolor{blue}{\textbf{6.62}}   \\\bottomrule
\end{tabular}}
\raggedright
\footnotesize{$^\dagger$ Expanded to 3 semantically equivalent prompts per test case.}  \\
\footnotesize{$^\ddagger$ Expanded to 5 semantically equivalent prompts per test case.}
\end{table}

\begin{figure}[!t]
\centering
        \begin{subfigure}{0.45\linewidth}
            \centering
            \includegraphics[width=0.95\linewidth]{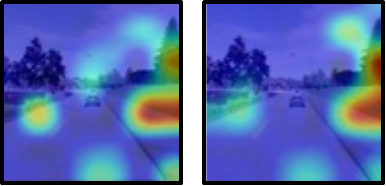}
            \caption{Attention Map before Attack}
            \label{fig:attention_lmdrive}
        \end{subfigure}
        \begin{subfigure}{0.45\linewidth}
            \centering
            \includegraphics[width=0.95\linewidth]{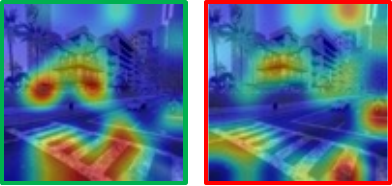}
            \caption{Attention Map after Attack}
            \label{fig:attention_lmdrive_new}
        \end{subfigure}
\caption{Results of the attention analysis.}
\vspace{-0.1in}
\label{fig:attention_new}

\end{figure}

\textbf{Model Attention Analysis.}
This section analyzes model attention through qualitative and quantitative studies to understand \toolns more thoroughly. Specifically, we examine attention maps from DriveLM and LMDrive, comparing them before and after the attack. Qualitatively, as shown in \Fref{fig:attention_lmdrive}, models initially focus on similar regions across prompts and perspectives, while after applying \toolns (see \Fref{fig:attention_lmdrive_new}), these attention maps shift significantly. Quantitatively, SSIM \cite{wang2004image_SSIM} and PCC metrics reveal high attention similarity across prompts and viewpoints before the attack (88.70\% and 88.27\% for DriveLM; 86.16\% and 90.83\% for LMDrive). Following the introduction of \toolns, these values drop significantly (to 26.74\% and 14.58\% for DriveLM; 37.45\% and 24.96\% for LMDrive), confirming that \tool disrupts stable attention patterns effectively. 

\section{Case Study for Real-World Attacks}
\label{sec: case}

In this section, we test our \tool on a real-world AD vehicle to further reveal the potential risks.

\begin{figure}[!t]
\centering
	\begin{subfigure}{0.45\linewidth}
		\centering
        \includegraphics[width=0.95\linewidth]{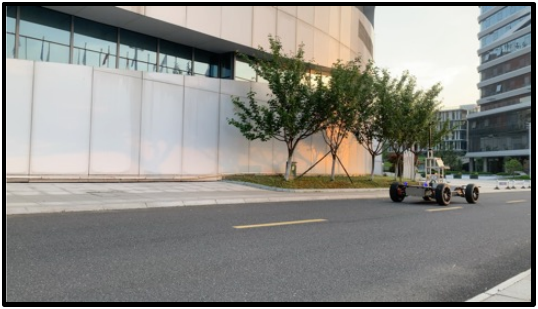}
		\caption{Environment}
		\label{subfig:real_world_1}
	\end{subfigure}
	\begin{subfigure}{0.45\linewidth}
		\centering
        \includegraphics[width=0.95\linewidth]{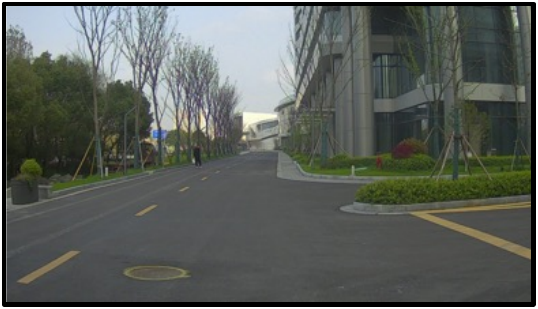}
		\caption{First-view Perspective}
		\label{subfig:real_world_2}
	\end{subfigure}
        \begin{subfigure}{0.45\linewidth}
		\centering
        \includegraphics[width=0.95\linewidth]{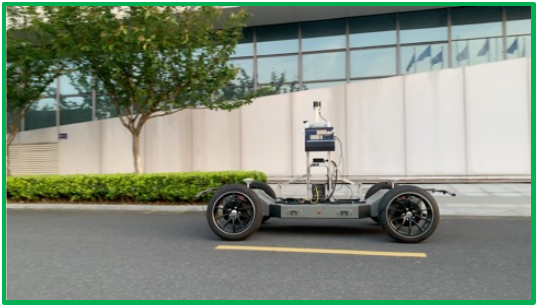}
		\caption{Driving without attack}
		\label{subfig:real_world_3}
	\end{subfigure}
        \begin{subfigure}{0.45\linewidth}
		\centering
        \includegraphics[width=0.95\linewidth]{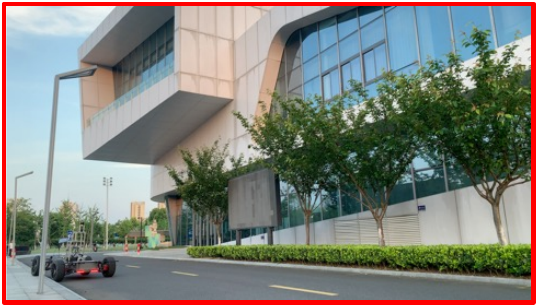}
		\caption{Driving with \tool attack}
		\label{subfig:real_world_4}
	\end{subfigure}
\caption{Real-World Case Study of \tool Attack. (a) Experimental environment setup. (b) First-person view from the vehicle. (c) Normal driving without attack, with the vehicle following the intended path. (d) \tool attack effect, causing the vehicle to deviate from its path, demonstrating potential real-world safety risks.}
\label{fig:real_world}
\vspace{-0.1in}
\end{figure}

\textbf{Experimental Setup.}
The experiment utilized a beta-version autonomous vehicle with a PIXLOOP-Hooke chassis \cite{pixloop}. This vehicle was outfitted with multiple perception and motion modules, including an RGB camera LI-USB30-AR023ZWDR to execute navigational commands provided by the VLM (\ie, Dolphins). We use high-level commands like ``go straight'' to translate into specific responses \texttt{drive\_mode\_ctrl} via the chassis. The prompt ``go straight'' was issued to the VLM. The environment and first-person view images displayed in \Fref{subfig:real_world_1} and \Fref{subfig:real_world_2}. Real-time adversarial noise generation by \tool was applied directly to the input, and the experiment was repeated 10 times both w/ and w/o attacks in daylight conditions.

\textbf{Results and Interpretation.}
In trials influenced by \toolns, the vehicle deviated from its intended route in 70\% of attempts, compared to 0\% in clean trials. Normal and deviated driving images are shown in \Fref{subfig:real_world_3} and \Fref{subfig:real_world_4}. Analysis of logged data packets indicated that under \tool’s attack, the RGB camera failed to capture critical road features, leading to off-course commands. Among the 7 successful attack-induced deviations, only 2 generated a warning and braking response within 0.5 seconds, significantly shorter than the average 2.5-second human reaction time \cite{sato2021dirty}. These findings highlight the tangible risks posed by \tool to real-world AD systems.

\section{Conclusion and Limitations}
\label{sec: conclusion}

This paper introduces \toolns, the first adversarial attack framework tailored specifically for VLMs in AD. \tool leverages Semantic-Invariant Induction within the textual domain and Scenario-Associated Enhancement within the visual domain, maintaining high attack effectiveness across diverse instructions and dynamic viewpoints. Extensive experiments demonstrate that \tool surpasses existing attack methods, highlighting substantial risks to AD systems.

\textbf{Limitations.} Despite the promising results, several areas remain to be explored: \ding{182} develop universal attack frameworks, \ding{183} explore targeted attack potential, and \ding{184} assess attack feasibility in additional or multimodal settings.

{
    \small
    \bibliographystyle{ieeenat_fullname}
    \bibliography{main}
}


\end{document}